\documentclass[journal]{IEEEtran}

\usepackage{amsmath,amssymb,mathtools}
\usepackage{graphicx}
\usepackage{booktabs}
\usepackage{tabularx}
\usepackage{longtable}
\usepackage{multirow}
\usepackage{array}
\usepackage{cite}

\usepackage{microtype}
\usepackage{algorithm}
\usepackage{algpseudocode}
\usepackage[dvipsnames]{xcolor}
\usepackage{caption}
\usepackage{subcaption}
\usepackage{enumitem}
\usepackage{footnote}
\usepackage{afterpage}

\usepackage{hyperref}
\hypersetup{
    colorlinks=true,
    linkcolor=Blue,
    filecolor=Magenta,
    urlcolor=Black,
    citecolor=Blue,
    pdftitle={StutterFuse: Mitigating Modality Collapse in Stuttering Detection with Jaccard-Weighted Metric Learning and Gated Fusion},
    pdfauthor={Guransh Singh},
}

\begin{document}

\title{StutterFuse: Mitigating Modality Collapse in Stuttering Detection with Jaccard-Weighted Metric Learning and Gated Fusion}

\author{Guransh Singh and Md. Shah Fahad%
\thanks{The authors are with the Department of Computer Science, Birla Institute of Technology, Mesra (e-mail: guransh766@gmail.com; fahad8siddiqui@bitmesra.ac.in).}}

\maketitle
\begin{abstract}
Stuttering detection breaks down when disfluencies overlap. Existing parametric models struggle to distinguish complex, simultaneous disfluencies (e.g., a 'block' with a 'prolongation') due to the scarcity of these specific combinations in training data. While Retrieval-Augmented Generation (RAG) has revolutionized NLP by grounding models in external knowledge, this paradigm remains unexplored in pathological speech processing.
To bridge this gap, we introduce \textit{StutterFuse}, the first Retrieval-Augmented Classifier (RAC) for multi-label stuttering detection. By conditioning a Conformer encoder on a non-parametric memory bank of clinical examples, we allow the model to classify by reference rather than memorization.
We further identify and solve "Modality Collapse"—an "Echo Chamber" effect where naive retrieval boosts recall but degrades precision. We mitigate this using: (1) \textit{SetCon}, a Jaccard-Weighted Metric Learning objective that optimizes for multi-label set similarity, and (2) a \textit{Gated Mixture-of-Experts} fusion strategy that dynamically arbitrates between acoustic evidence and retrieved context.
On the SEP-28k dataset, StutterFuse achieves a weighted F1-score of \textit{0.65}, outperforming strong baselines and demonstrating remarkable zero-shot cross-lingual generalization.
\end{abstract}


\section{Introduction}
\label{sec:introduction}

Current speech recognition models fail when faced with the irregular rhythmic interruptions typical of stuttering—specifically blocks, prolongations, and repetitions. While often dismissed as simple repetitions, stuttering is a complex neurodevelopmental condition that breaks the flow and natural rhythm of speech. It isn't just one thing; it's a collection of disfluencies like sound repetitions (\textit{S-S-SoundRep}), word repetitions (\textit{Word-WordRep}), and blocks, where airflow hits a wall \cite{dsm5stuttering}. Beyond those things, stuttering carries a heavy weight, often affecting communication confidence and personal life from relationships to career paths \cite{yairi2013epidemiology}.

Automated detection tools are critical for addressing these issues. They provide objective metrics for speech-language pathologists (SLPs) and real-time feedback for people who stutter (PWS), while enabling more inclusive human-computer interaction.

Building these systems is difficult because stuttering data is messy. A single clip might contain multiple overlapping disfluencies. This creates a combinatorial explosion of labels that standard classifiers can't handle. Worse, the acoustic signs are subtle and vary wildly between speakers, a problem compounded by the scarcity of high-quality, annotated datasets. Beacuse of these issues, simple classifiers often fail.

Prior work has often fallen short of addressing this complexity. Many studies resort to binary (stuttered vs. fluent) classification, use speaker-overlapping data splits that lead to artificially inflated results, or treat the task as single-label, forcing a "winner-takes-all" choice that does not reflect clinical reality.

In this work, we tackle the problem head-on by formulating stuttering detection as a rigorous, speaker-independent, multi-label classification task. We demonstrate that even high-capacity, modern architectures like the Conformer \cite{gulati2020conformer}, when trained conventionally, struggle with the combinatorial explosion of label sets and the imbalance of rare combinations. By leveraging a retrieval-based approach, we aim to bypass the limitations of "parametric-only" learning, allowing the model to reference specific, stored examples of these rare combinations during inference.

Our central hypothesis is that a conventional classifier can be significantly improved by explicitly providing it with \textit{retrieved reference examples} from the most similar, already-labeled instances in the training set. We propose \textit{StutterFuse}, a retrieval-augmented pipeline that learns to "classify by non-local comparison." Our contributions are:-
\begin{enumerate}
    \item \textit{Retrieval-Augmented Architecture:} We built \textit{StutterFuse}, a system that pairs a standard classifier with a memory bank. By adding a \textit{Late-Fusion (Gated Experts)} mechanism, we solved the "Echo Chamber" problem—where the model essentially copies its neighbors—allowing it to balance acoustic evidence with retrieved context dynamically.
    \item \textit{Jaccard-Weighted Metric Learning:} Standard losses like Multi-Similarity didn't work well for our multi-label data. So, we designed \textit{Set-Similarity Contrastive Loss (SetCon)}. This loss function cares about the overlap of the entire label set (Jaccard similarity), which bumped our retrieval recall from 0.32 up to 0.47.
    \item \textit{Cross-Attention Fusion:} In our Mid-Fusion baseline, we used a cross-attention module that doesn't just look at neighbor embeddings. It looks at a "value" matrix containing their acoustics, labels, and similarity scores, letting the model decide exactly which pieces of information to trust.
    \item \textit{Reproducible Results:} We are releasing the full pipeline, which achieves a weighted F1-score of 0.65. While perfectly clean detection remains elusive, this score is approaching the limit of human agreement (Kappa $\approx$ 0.7) \cite{lea2021sep28k}, and we provide a full breakdown of why (Section \ref{sec:ablations}).
\end{enumerate}

We are open-sourcing the embedder, Faiss index, and training scripts to give the community a solid, reproducible starting point: \url{https://github.com/GS-GOAT/Stutter-Speech-Classifier/}.

\section{Related Work}
\label{sec:related_work}
Our work sits at the intersection of three primary research areas: (1) automatic stuttering detection, (2) advances in deep learning for speech, and (3) the emerging paradigm of retrieval-augmented models.

\subsection{Classical and Early Deep Learning for Stuttering Detection}
The first generation of stuttering detection relied heavily on manual feature engineering. Researchers would extract MFCCs, LPC coefficients, or prosodic features (pitch, energy) and feed them into standard classifiers like SVMs or GMMs \cite{vinod2015mfcc}. These methods were foundational but brittle; they often failed to generalize across different speakers or capture the wide variety of disfluency shapes.

Deep learning shifted the paradigm by automating feature extraction. Early neural approaches—ranging from basic DNNs \cite{ma2018dnn} to CNNs \cite{chee2016dnn} and RNNs \cite{einarsdottir2019stuttering}—showed they could learn complex, hierarchical patterns directly from raw audio or spectrograms. They comfortably outperformed the older feature-based models \cite{kourkounakis2020fluentnet}. But they hit a wall: deep models are data-hungry, and in the niche field of pathological speech, labeled data is scarce.

\subsection{Self-Supervised Learning in Speech Processing}
The recent breakthrough in speech processing has been the development of self-supervised learning (SSL) models, like Wav2Vec 2.0 \cite{baevski2020wav2vec}, HuBERT \cite{hsu2021hubert}, and data2vec \cite{baevski2022data2vec} learn high quality contextualized representations from large amounts of unlabeled audio. These foundation models for speech can then be fine-tuned on downstream tasks with limited labeled data, or used as-is as high quality "frozen" feature extractors. The Wav2Vec 2.0 model, in particular, uses a contrastive loss to learn discrete speech units, making it highly effective at capturing phonetic and sub-phonetic details that are important for identifying the subtle acoustic anomalies associated with stuttering.

This approach has been successfully applied to stuttering. \cite{takashima2022wav2vecstutter} showed that fine-tuning a Wav2Vec 2.0 model yields state-of-the-art results in stuttering detection. Our work builds on this, using a frozen Wav2Vec 2.0 as a high-quality, reproducible feature extractor. This allows us to decouple the feature representation from the main contribution of our work: the retrieval-augmented architecture built on top of these features. By using frozen features, we also significantly reduce the computational cost of training our subsequent metric learning and classifier stages.

\subsection{Advanced Architectures: Attention and Conformers}
Within the deep learning paradigm, attention mechanisms and the Transformer architecture revolutionized sequence modeling. For stuttering, attention-based BiLSTMs were explored in models like StutterNet \cite{sheikh2021stutternet}, which showed an ability to focus on salient parts of the acoustic signal.

More recently, the Conformer architecture \cite{gulati2020conformer} has become the \textit{de facto} state-of-the-art for most speech tasks. By effectively combining the local feature extraction of CNNs (via depthwise-separable convolutions) with the global context modeling of Transformers \cite{vaswani2017attention} (via multi-head self-attention), Conformers provide an excellent inductive bias for speech. This motivates our choice of a Conformer-based architecture as both our strong baseline and the encoder for our final StutterFuse model.

\subsection{Metric Learning and Retrieval-Augmentation}
Standard parametric models are fundamentally limited by their reliance on fixed weights to encode the entire variation of stuttering phenomenology, often failing to capture the long tail of rare and complex disfluencies. To overcome this "memorization bottleneck", we draw upon two complementary paradigms: Metric Learning, which organizes the latent space to reflect clinical similarity rather than just acoustic proximity, and Retrieval-Augmentation, which allows the model to dynamically reference explicit examples from a memory bank, thereby bridging the gap between specific instance recall and general pattern matching.

\paragraph{Metric Learning} seeks to learn an embedding space where a chosen similarity metric (e.g., Euclidean distance) corresponds to a desired semantic similarity. This is often achieved via siamese networks or, like, triplet loss \cite{hoffer2015triplet}. Triplet loss, famously used in FaceNet \cite{schroff2015facenet} and audio retrieval \cite{jansen2018unsupervised}, trains a model by optimizing for a ``margin'' where an anchor sample is closer to positive samples (same class) than to negative samples (different class) \cite{hermans2017defense}. While common in vision, its application to complex, multi-label audio events like stuttering is less explored.

\paragraph{Retrieval-Augmentation} is a paradigm, primarily applied in large language models (e.g., RAG \cite{lewis2020rag}), where a model's prediction is explicitly conditioned on information retrieved from a large, external database. This has the effect of \textit{bolting on} a massive memory to the model. This concept is not new—the k-Nearest Neighbors (k-NN) algorithm is its simplest form. Our work builds a modern, deep-learning-native version of this: we use a learned metric space for retrieval (not just raw features) and a deep-learning-based fusion mechanism (cross-attention) to integrate the retrieved information, rather than a simple majority-vote. This hybrid, "parametic + non-parametric" approach is what defines StutterFuse.

\section{Dataset and Problem Formulation}
\label{sec:dataset}

In this section, we detail the data and mathematical framework used in our study. We first describe the primary training corpus (SEP-28k) and the two out-of-domain evaluation datasets (FluencyBank and KSOF) used to test robustness. We then analyze the label distribution and our preprocessing strategies for handling class imbalance, before formally defining the multi-label classification task and the Jaccard-based similarity metric.

\subsection{SEP-28k Dataset}
Our primary dataset is SEP-28k \cite{lea2021sep28k}, one of the largest publicly available corpora for stuttering detection. It consists of approximately 28,000 audio clips, each roughly 3 seconds in duration, sourced from podcasts featuring PWS. The clips are annotated with a multi-label ontology derived from the Stuttering-Severity-Instrument 4 (SSI-4), including \texttt{Prolongation}, \texttt{Block}, \texttt{SoundRep}, \texttt{WordRep}, \texttt{Interjection}, and \texttt{NoStutter}.

A vital feature of this dataset is the availability of speaker/show identifiers, which is essential for creating clinically-relevant, speaker-independent evaluation splits. We rigorously enforce this, ensuring that no speaker in the test set is present in the training or validation sets.

\subsection{FluencyBank Dataset}
To assess the out-of-domain generalization of our models, we use the FluencyBank dataset as a secondary test set \cite{howell2009fluencybank}. FluencyBank consists of speech from PWS in more structured, clinical, or conversational settings, differing significantly from the podcast-style audio of SEP-28k. The labels were mapped from FluencyBank's scheme to the SEP-28k ontology by \cite{bayerl2023}. Evaluating on FluencyBank provides a robust test of whether our model has learned fundamental acoustic properties of disfluencies or has simply overfit to the acoustic environment of SEP-28k.

\subsection{KSOF Dataset}
We also utilize the Kassel State of Fluency (KSoF) dataset \cite{bayerl2022ksof} to investigate cross-lingual generalization. The KSoF is a German stuttering dataset sourced from various media formats. Although the language differs, the physiological manifestations of core stuttering events (like blocks and prolongations) share acoustic similarities with English. We use the test split of KSoF and map its labels to the SEP-28k ontology to perform a zero-shot evaluation of our English-trained models.

\subsection{Label Distribution and Data Preprocessing}
\label{ssec:data_preprocessing}
As our goal is to characterize \textit{which} disfluencies are present in a stuttered segment, we first filter the dataset to remove all clips labeled \textit{only} as \texttt{NoStutter}. We focused on careful data augmentation strategies, as in a multi-label setting, naively upsampling minority classes can inadvertently increase the distribution of majority classes if they co-occur, necessitating a nuanced approach.

The resulting label distribution in our speaker-independent training split (post-filtering) is shown in Table \ref{tab:label_dist}. The dataset exhibits a severe long-tail imbalance: \texttt{Block} is present in 8,081 clips, whereas \texttt{WordRep} is present in only 2,759. This imbalance poses a significant challenge, as a naive classifier will be heavily biased towards the majority classes.

\begin{table*}[!t]
\centering
\caption{Label counts in the speaker-aware training set before and after instance-balanced augmentation.}
\label{tab:label_dist}
\begin{tabular}{lrr}
\toprule
\textit{Disfluency Class} & \textit{Original Count} & \textit{Augmented Count} \\
\midrule
Block & 8081 & 10,848 \\
Interjection & 5934 & 10,824 \\
Prolongation & 5629 & 10,800 \\
SoundRep & 3486 & 10,576 \\
WordRep & 2759 & 10,569 \\
\bottomrule
\end{tabular}

\end{table*}

\begin{figure}[ht]
  \centering
  \includegraphics[width=\linewidth]{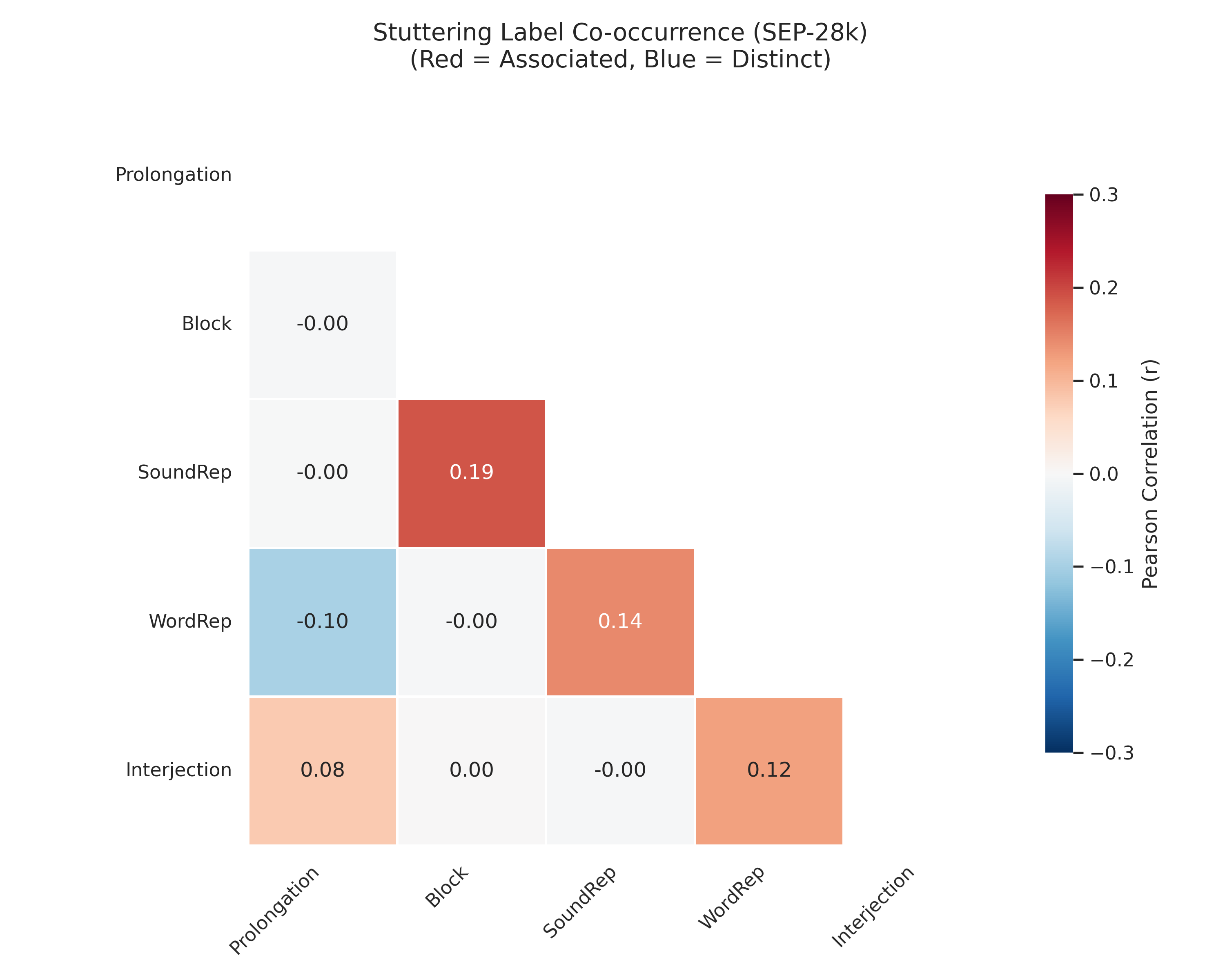}
  \caption{Pearson correlation matrix of disfluency labels in the SEP-28k training set. The low correlation values indicate that stuttering types are largely independent, supporting the formulation of the task as a multi-label problem rather than multi-class.}
  \label{fig:cooccurrence}
\end{figure}

Furthermore, the multi-label nature of the data introduces specific co-occurrence patterns, as shown in the Pearson correlation heatmap in Figure \ref{fig:cooccurrence}. While most disfluency types are statistically distinct (near-zero correlation), we observe weak positive correlations between struggle behaviors (e.g., \texttt{Block} and \texttt{SoundRep}, r=0.19) and negative correlations between distinct types (e.g., \texttt{WordRep} and \texttt{Prolongation}, r=-0.10). This is a critical challenge: a simple "instance-balanced" augmentation (described in \ref{ssec:augmentation}) that duplicates a \texttt{WordRep} clip may also duplicate a co-occurring \texttt{Block}, which can inadvertently worsen the majority-class bias. Our metric learning stage (Section \ref{ssec:metric_learning}) is designed specifically to handle this combinatorial complexity.

\subsection{Problem Formulation}
We formulate the task as multi-label classification.
\begin{itemize}
    \item Let $X$ be a 3-second raw audio clip, which we represent as a feature sequence $\mathbf{F}_X \in \mathbb{R}^{T \times D}$, where $T=150$ frames and $D=1024$ features (from Wav2Vec 2.0).
    \item Let $C = 5$ be the number of disfluency classes.
    \item The target is a binary vector $y \in \{0, 1\}^C$, where $y_i = 1$ if the $i$-th disfluency is present, and $y_i = 0$ otherwise.
    \item The model $M$ must learn a mapping $M(\mathbf{F}_X) \to \hat{y}$, where $\hat{y} \in [0, 1]^C$ is a vector of predicted probabilities, typically from a sigmoid activation.
\end{itemize}

For our metric learning stage, we define a label similarity metric. We use the \textit{Jaccard distance} $d_J(y_a, y_b)$ between two label vectors $y_a$ and $y_b$, which is defined as:
\[
d_J(y_a, y_b) = 1 - \frac{|y_a \cap y_b|}{|y_a \cup y_b|} = \frac{|(y_a \neq y_b) \land (y_a \lor y_b)|}{|y_a \lor y_b|}
\]
This metric is 0 for identical label vectors and 1 for perfectly disjoint, non-empty vectors. It is the ideal metric for our similarity-based mining, as it directly measures the similarity of the multi-label sets.

\section{Methodology: The StutterFuse Pipeline}
\label{sec:methodology}

\begin{figure*}[!t]
  \centering
  \includegraphics[width=1.0\textwidth]{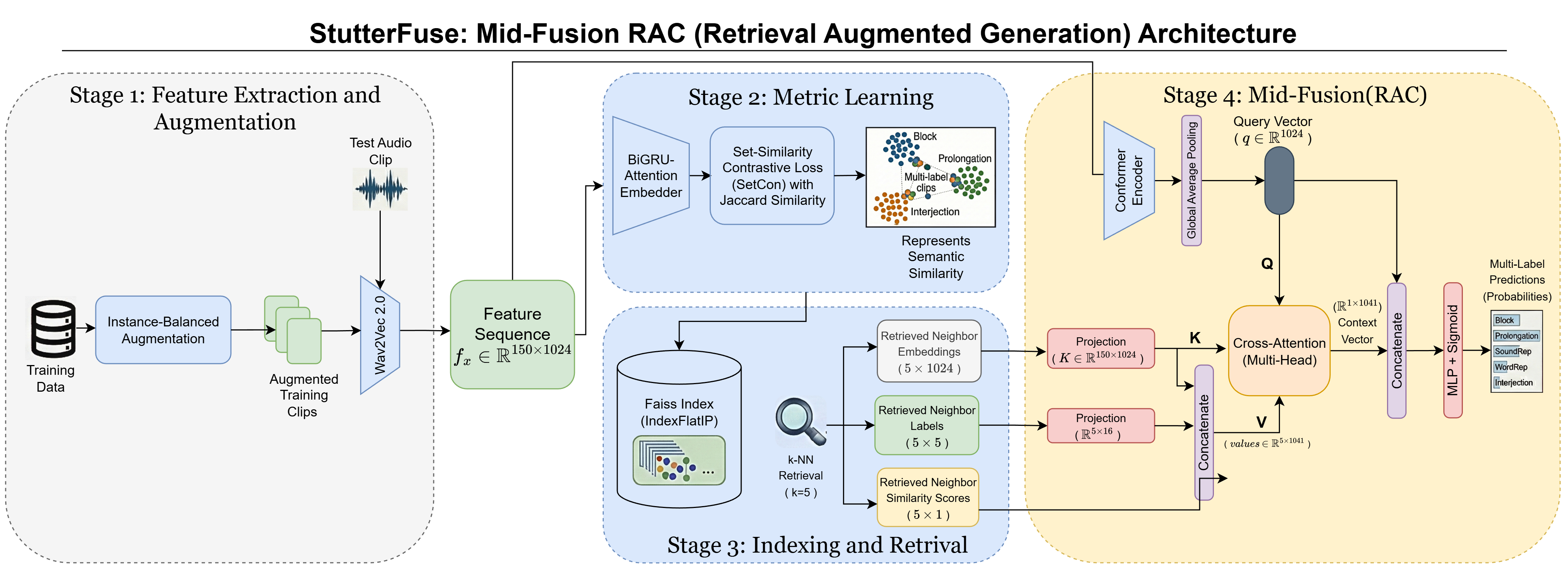}

  \caption{A high-level overview of the proposed StutterFuse pipeline. (1) A specialized embedder is trained with Multi-Similarity Loss. (2) A Faiss index is built from these embeddings. (3) The final classifier uses cross-attention to fuse information from the query clip and its $k$-retrieved neighbors.}
  \label{fig:pipeline_overview}
\end{figure*}

Our proposed pipeline, StutterFuse, is a multi-stage process designed to tackle the challenges of multi-label classification and class imbalance through retrieval-augmentation. The full pipeline is illustrated in Figure \ref{fig:pipeline_overview}.

\subsection{Stage 1: Feature Extraction and Augmentation}
\label{ssec:augmentation}

\paragraph{4.1.1. Wav2Vec 2.0 Feature Extraction}
We use the \texttt{facebook/wav2vec2-large-960h} model as a frozen feature extractor. Each 3-second audio clip is passed through the model, and we extract the hidden states from its final transformer layer. This sequence is center-truncated or padded to a fixed length of $T=150$ frames. This results in a feature tensor $\mathbf{F}_X \in \mathbb{R}^{150 \times 1024}$ for each clip. Precomputing these features Significantly reduces training time and ensures a consistent, high-quality input for all subsequent experiments.

\paragraph{4.1.2. Instance-Balanced Augmentation via Optimization}
As Table \ref{tab:label_dist} highlights, the raw training set is heavily skewed. Traditional class-wise oversampling doesn't work well here; if you duplicate a clip to boost a rare label like \texttt{WordRep}, you often unintentionally boost a co-occurring common label like \texttt{Block}, which just distorts the marginal distributions further.

To get around this, we treat augmentation as an optimization problem. For every class $c$, we calculate a rarity score $r_c = 1/f_c$, based on raw frequency $f_c$. For any clip $x$ with a set of labels $L_x$, we define its instance rarity as $r_x = \min_{c \in L_x} r_c$.

Rather than deterministically augmenting every instance, we formulate the selection of augmentation candidates as a constrained optimization problem. Let $a_{i,k} \in \{0, 1\}$ indicate whether instance $i$ receives its $k$-th augmentation copy. We aim to minimize the weighted squared error between the final class counts $\hat{F}_c$ and target counts $T_c$:
\[
\min_{a_{i,k}} \sum_{c} w_c (\hat{F}_c - T_c)^2
\]
subject to per-instance caps based on rarity ($ \sum_k a_{i,k} \le \lceil g(r_x) \rceil $) and a total budget constraint. We solve this using a greedy heuristic that iteratively selects the instance-copy yielding the greatest reduction in the objective.

We use \texttt{audiomentations} (GaussianNoise, PitchShift, TimeStretch) to generate the selected copies. This process yields the near-equalized distribution detailed in Table \ref{tab:label_dist}, providing sufficient minority-class exposure without massive co-occurrence distortion.

\subsection{Stage 2: Metric Learning for Stuttering Similarity}
\label{ssec:metric_learning}
A simple average-pooling of the Wav2Vec 2.0 features is not optimized for retrieval based on \textit{label similarity}. Our key insight is to train a new, dedicated embedder for this purpose.

\paragraph{4.2.1. Embedding Architecture}
We design a specialized "BiGRU-Attention" embedder to transform the $150 \times 1024$ sequence into a single, fixed-size vector $\mathbf{e} \in \mathbb{R}^{1024}$. The architecture, detailed in \texttt{embedr.py}, consists of:
\begin{enumerate}
    \item \textit{BiGRU Encoder:} We employ a Bidirectional Gated Recurrent Unit (BiGRU) as the primary sequence encoder to capture the temporal dynamics of disfluencies. With 256 hidden units per direction, it processes the 150-frame Wav2Vec 2.0 sequence, producing a 512-dimensional output at each time step. This bidirectional processing ensures that the representation of any potential stuttering event is informed by both preceding and succeeding acoustic context, capturing the full envelope of the disfluency.
    \item \textit{Projection:} A dense projection layer expands the encoder's output to a higher-dimensional space of 1024 units, utilizing a ReLU activation. Intentionally avoiding bottlenecks, this high-dimensional projection preserves the rich information capacity needed to represent complex, overlapping multi-label combinations. This allows the network to "untangle" the feature manifold before the aggregation step, ensuring that subtle acoustic cues distinguishing rare classes are not compressed away.
    \item \textit{Attention Pooling:} Instead of rigid global averaging, we implement a learnable self-attention mechanism to aggregate the temporal sequence. A dedicated scoring head (Dense $\to$ Tanh $\to$ Softmax) assigns a relevance weight to each frame, effectively learning to identify and prioritize the moments where stuttering occurs. The final pooled vector is computed as the weighted sum of the projected features, allowing the model to suppress silence or fluent segments and focus entirely on the pathological event.
    \item \textit{L2 Normalization:} The final aggregated embedding is passed through an L2 normalization layer, projecting the vector onto the unit hypersphere. This step is mathematically essential for our metric learning framework, as it makes the dot product equivalent to Cosine Similarity. By removing magnitude variations, we ensure that the subsequent neighbor retrieval is based purely on the semantic orientation of the embeddings, stabilizing the optimization of the contrastive loss.
\end{enumerate}

\paragraph{4.2.2. Set-Similarity Contrastive Loss (SetCon)}
We train the embedder using a \textit{SetCon} ($\tau=0.1$). Unlike standard Triplet Loss, SetCon leverages the full batch for contrastive learning and weighs positive pairs based on their label overlap.
\begin{equation}
    \mathcal{L} = \sum_{i \in I} \frac{-1}{|P(i)|} \sum_{p \in P(i)} w_{ip} \cdot \log \frac{\exp(z_i \cdot z_p / \tau)}{\sum_{a \in A(i)} \exp(z_i \cdot z_a / \tau)}
\end{equation}
The weight $w_{ip}$ is derived from the Jaccard similarity between the label sets of anchor $i$ and positive $p$. This approach resonates with recent findings in the NLP domain, such as the Jaccard Similarity Contrastive Loss (JSCL) proposed by \cite{lin2023jscl} for multi-label text classification. We extend this concept to the acoustic domain, ensuring that the model learns to cluster samples with similar \textit{multi-label profiles}, not just shared single labels.
\paragraph{Impact of SetCon and Failure of Standard Losses:}
We found that this Jaccard-weighted objective was critical for performance. We compared our approach against several strong baselines. The simple averaging of Wav2Vec 2.0 features (\textit{Mean Pooling}) yielded a Recall@5 of \textit{0.32}. State-of-the-art metric learning objectives provided moderate gains, with \textit{Multi-Similarity (MS) Loss} achieving \textit{0.39} and \textit{Standard Triplet Loss} reaching \textit{0.42}. However, our proposed \textit{SetCon} objective outperformed these methods significantly, achieving a Recall@5 of \textit{0.47}, confirming the importance of Jaccard-based optimization for multi-label retrieval.
Standard losses like MS Loss and SupCon are designed for \textit{multi-class} problems: they treat any shared label as a perfect match (binary positive) and push everything else away. In the multi-label stuttering domain, this is flawed. A "Block" and a "Block+WordRep" are similar but not identical. SetCon's continuous Jaccard weighting captures these nuances, creating a semantically rich space where partial overlaps are respected. This 16\% absolute gain over the baseline provides a far denser memory bank for the subsequent fusion stages.

\paragraph{4.2.3. Training and Validation}
We trained the SetCon model for 20 epochs using the Adam optimizer \cite{kingma2014adam} with a learning rate of 1e-4. The goal of this stage wasn't just to lower loss, but to improve retrieval quality. To track this, we built a custom \texttt{RecallAtKEvaluator}. At the end of every epoch, this callback would generate embeddings for the validation set, build a temporary index, and check if the top 10 neighbors were actually relevant (Set Jaccard Distance $\le 0.5$). The logs showed steady progress, eventually hitting a \textit{Recall@5 of 0.47}. This metric confirmed that our embedding space was actually learning semantic similarity. We saved the weights from the epoch that maximized this recall.

\begin{figure*}[!t]
  \centering
  \includegraphics[width=1.0\textwidth]{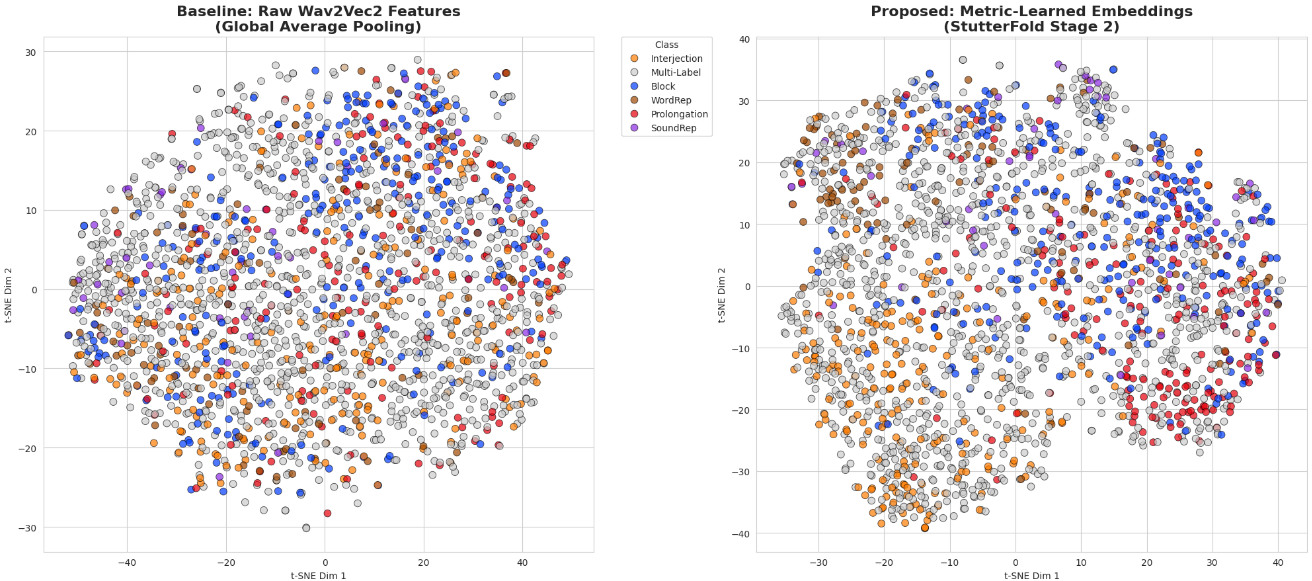}
  \caption{t-SNE visualization of the embedding space. Left: Raw Wav2Vec 2.0 features show poor separation. Right: Our metric-learned embeddings show distinct clusters for different stuttering types, facilitating effective retrieval. Note the formation of distinct clusters for Block and Prolongation, which were entangled in the Baseline space.}
  \label{fig:tsne_comparison}
\end{figure*}

\begin{figure}[!t]
  \centering
  \includegraphics[width=\columnwidth]{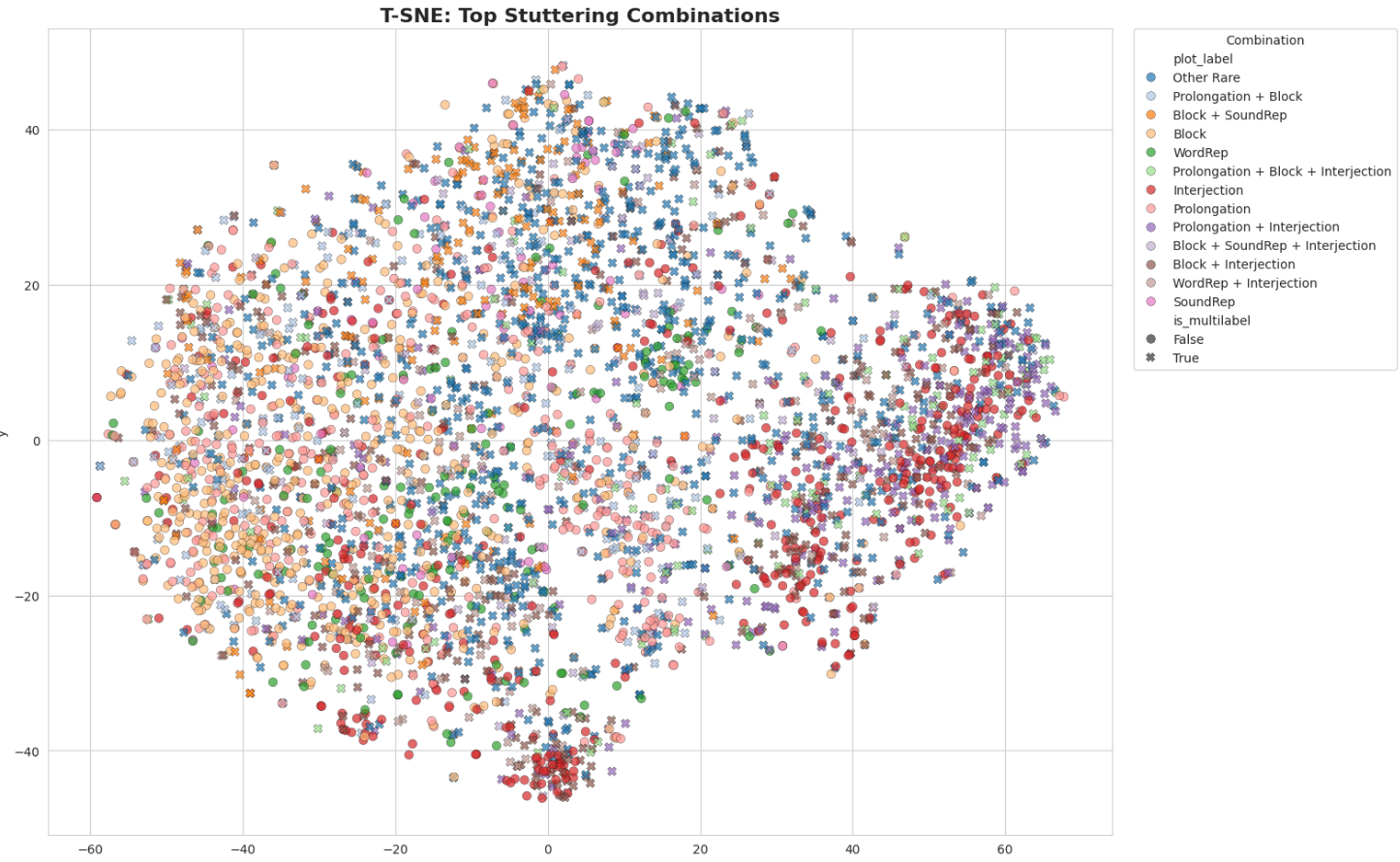}
  \caption{Interaction Map of the learned embedding space. The visualization reveals how multi-label combinations (e.g., Block + WordRep) cluster naturally between their constituent pure classes, validating the model's ability to capture compositional semantics.}
  \label{fig:tsne_interaction}
\end{figure}

\begin{figure*}[!t]
  \centering
  \includegraphics[width=1.0\textwidth]{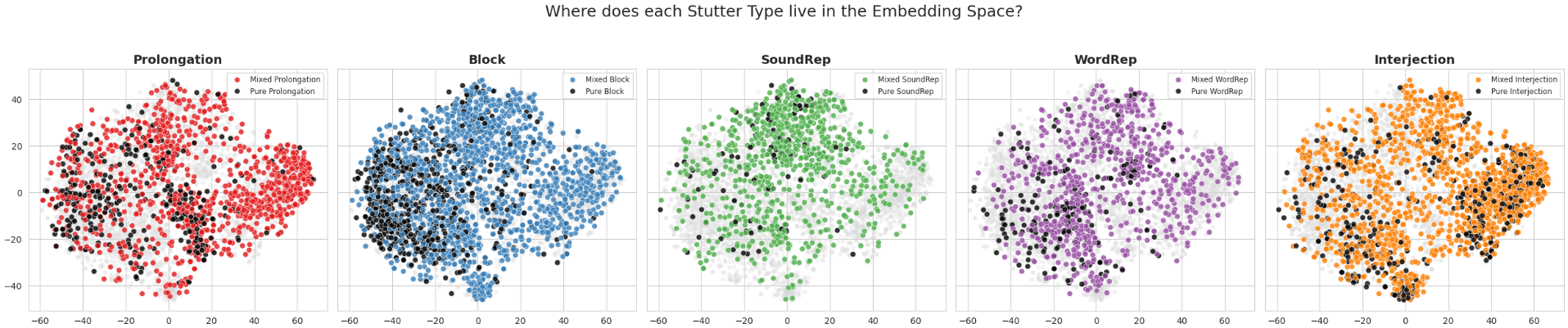}
  \caption{Class Decomposition Grid. We decompose the embedding space by class to analyze the distribution of each disfluency type. While 'Blocks' and 'Prolongations' form tight, distinct clusters, 'Interjections' are more diffuse, highlighting the acoustic variability of this class.}
  \label{fig:tsne_decomposition}
\end{figure*}

To double-check this visually, we projected the embeddings using t-SNE (Figure \ref{fig:tsne_comparison}). The difference is stark: while baseline Wav2Vec 2.0 features are a tangled mess, our learned embeddings form clear, distinct clusters for different stutter types. We further analyze the structure of this space in Figure \ref{fig:tsne_interaction}, which shows how multi-label combinations naturally cluster between their constituent pure classes, and Figure \ref{fig:tsne_decomposition}, which decomposes the space by class to highlight the compactness of Blocks versus the variance of Interjections.

\subsection{Stage 3: Retrieval-Augmented Classifier (RAC)}
\label{ssec:rac_model}
This is our final StutterFuse model, which uses the embedder from Stage 2 to perform classification.

\paragraph{4.3.1. ANN Index Construction}
We first use the trained and saved embedder model to compute the final, fixed-size embeddings $\mathbf{e} \in \mathbb{R}^{1024}$ for all clips in the augmented training set. These vectors are L2-normalized and stored in a \textit{Faiss} \texttt{IndexFlatIP} (Flat Inner Product) index. This index allows for efficient MIPS (Maximum Inner Product Search), which for L2-normalized vectors is equivalent to finding the nearest neighbors by cosine similarity.

\paragraph{4.3.2. Model Inputs}
The StutterFuse classifier is a multi-input model. For each batch sample, the data pipeline retrieves its $k=5$ nearest neighbors from the Faiss index and provides the model with the following dictionary of tensors:
\begin{enumerate}
    \item \texttt{input\_test\_seq}: The $T \times D$ Wav2Vec 2.0 sequence of the target clip (query).
    \item \texttt{input\_neighbor\_vecs}: The $k \times D$ pooled embedding vectors of its $k$ neighbors.
    \item \texttt{input\_neighbor\_labels}: The $k \times C$ ground-truth label vectors of the $k$ neighbors.
    \item \texttt{input\_neighbor\_sims}: The $k$ similarity scores (distances) from the Faiss search.
\end{enumerate}

\subsection{Phase 2: Fusion Strategies}
We explore two distinct strategies for integrating the retrieved information. Note that in both strategies, the retrieved neighbors are processed as \textit{vectors} (pooled embeddings), not full sequences, to maintain computational efficiency.

\subsubsection{Strategy A: Mid-Fusion via Cross-Attention (RAC)}
In this approach, we fuse the acoustic and retrieval streams \textit{before} the final classification head.
In this architecture, the \textbf{Query ($Q$)} is derived from the target audio, processed by a shared Conformer encoder and pooled to a vector $q \in \mathbb{R}^{1024}$. The \textbf{Keys ($K$) \& Values ($V$)} are formed from the $k$ retrieved neighbor vectors, which are projected to 1024 dimensions to create the Keys. The Values consist of a concatenation of these Keys, the neighbor label embeddings (Dense 16), and their similarity scores. Finally, a \textbf{Cross-Attention} layer (4 heads, key\_dim=256) attends to these neighbors, computing the context as $\text{Context} = \text{Attention}(Q=q, K=K_{\text{neigh}}, V=V_{\text{neigh}})$.
The context vector is concatenated with the query and passed to the final MLP (Dense 512 $\to$ Dropout $\to$ Dense 256 $\to$ Output).

\subsubsection{Strategy B: Late-Fusion via Gated Experts (StutterFuse)}
\begin{figure*}[!t]
  \centering
  \includegraphics[width=1.0\textwidth]{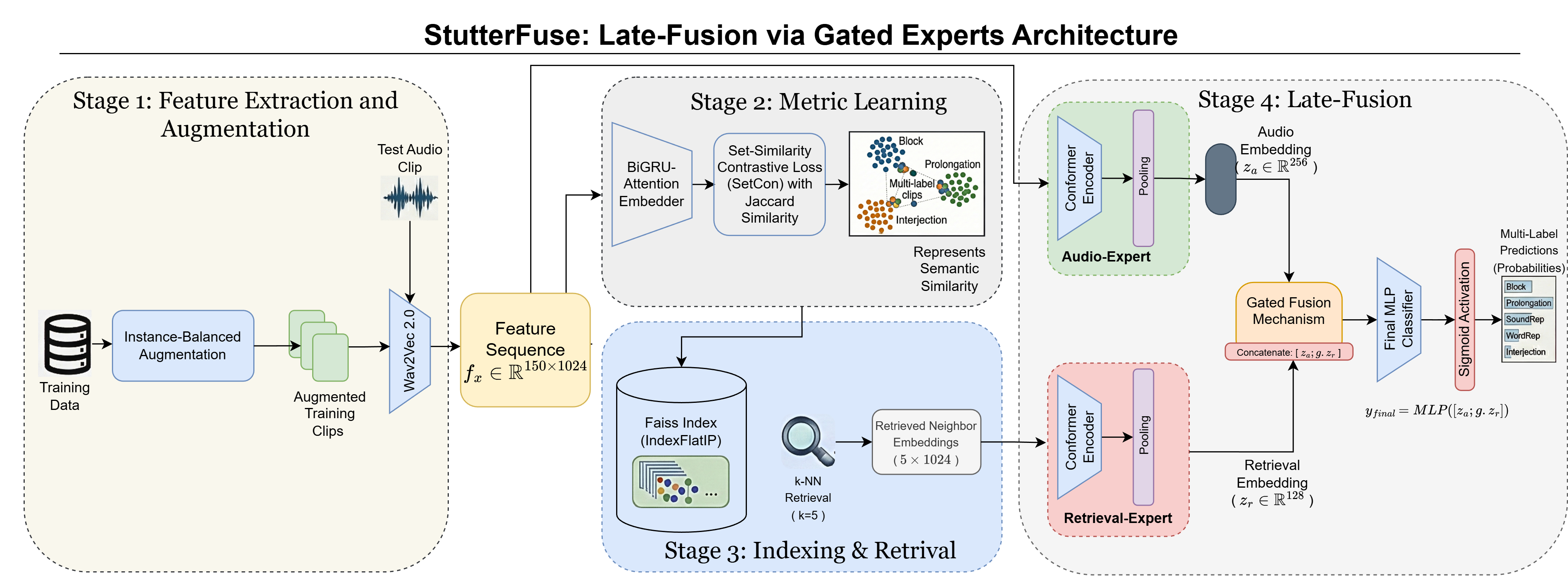}
  \caption{The StutterFuse Architecture (Late-Fusion). The model processes the audio and retrieved neighbors via separate expert streams. A learned gate dynamically weighs the retrieval expert's contribution before the final classification.}
  \label{fig:stutterfuse_arch}
\end{figure*}

To address the "Echo Chamber" effect, we propose a \textit{Late-Fusion} architecture that treats the streams as independent experts.
\begin{enumerate}
    \item \textit{Expert A (Audio):} A Conformer (2 blocks, ff\_dim=512) processes the target audio. Output: $z_a \in \mathbb{R}^{256}$.
    \item \textit{Expert B (Retrieval):} A lightweight MLP processes the $k$ neighbor vectors (Dense 256 $\to$ GlobalAvgPool $\to$ Dense 128). Output: $z_r \in \mathbb{R}^{128}$.
    \item \textit{Gated Fusion:} A learned gate $g \in [0, 1]$ dynamically weighs the retrieval expert:
    \begin{equation}
        g = \sigma(W_g [z_a; z_r] + b_g), \quad y_{final} = \text{MLP}([z_a; g \cdot z_r])
    \end{equation}
\end{enumerate}
This \textit{StutterFuse} model (Figure \ref{fig:stutterfuse_arch}) allows the network to suppress the retrieval stream when the acoustic signal is unambiguous.

\section{Experimental Setup}
\label{sec:experimental_setup}

\subsection{Implementation Details}
All models were implemented in TensorFlow 2.x with Keras. We used the \texttt{faiss-cpu} library \cite{johnson2019billion} for nearest-neighbor search. Due to the large model size and batch requirements, all experiments were conducted on a Google Cloud TPU v5e-8, using \texttt{tf.distribute.TPUStrategy} for distributed training. The global batch size was set to 128 (16 per replica).

\begin{figure*}[!t]
  \centering
  \includegraphics[width=0.8\textwidth]{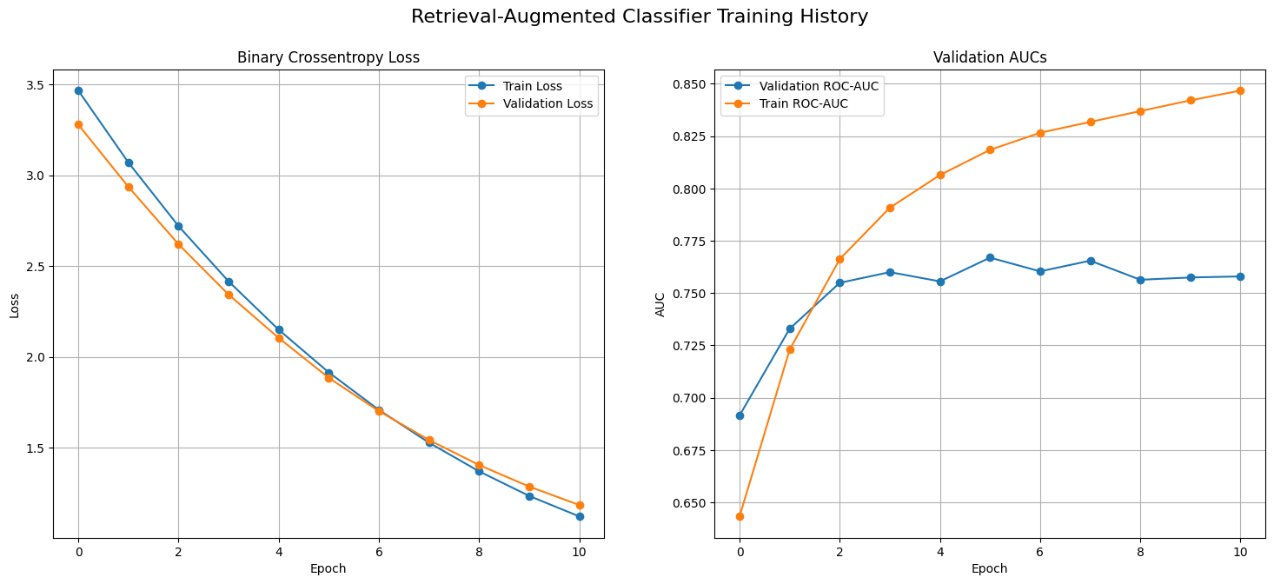}
  \caption{Training dynamics of the StutterFuse model. Left: Binary Cross-Entropy Loss over epochs. Right: Validation AUC-ROC score, demonstrating stable convergence.}
  \label{fig:training_curves}
\end{figure*}

\paragraph{Stage 2 (Embedder) Training:} Adam optimizer, LR=1e-4. Trained for 20 epochs, with the best model selected by \texttt{val\_recall\_at\_5}.
\paragraph{Stage 3 (RAC) Training:} AdamW optimizer, LR=2e-5, weight decay=5e-4. Loss was Binary Cross-Entropy with 0.1 label smoothing. We used \texttt{val\_auc\_roc} as the monitoring metric for early stopping with a patience of 5, as it is a threshold-independent metric suitable for imbalanced data. Alternative losses like Focal Loss \cite{lin2017focal} were considered but BCE with smoothing proved more stable. Figure \ref{fig:training_curves} illustrates the training progression, showing the convergence of the loss and the steady improvement of the AUC-ROC score.

\subsection{Baselines for Comparison}
To quantify the benefit of our pipeline, we compare StutterFuse against a suite of strong baselines, all trained on the same frozen Wav2Vec 2.0 features and instance-balanced augmented data:
We initially experimented with several standard deep learning architectures, including a \textit{DNN}, \textit{CNN}, and a standard \textit{Transformer} encoder. However, preliminary observations indicated that these models were less effective at capturing the subtle, multi-scale dynamics of stuttering. Consequently, we chose the \textit{Conformer (No Retrieval)} as our primary baseline, as it provided the strongest parametric performance. This model consists of a 2-block Conformer encoder (4 heads, ff\_dim=1028). Unlike standard approaches that pool features before classification, this model applies a TimeDistributed Dense layer with sigmoid activation to the frame sequence, followed by Global Average Pooling, ensuring it can detect disfluencies occurring at any point in the clip.

\subsection{Evaluation Metrics}
Since this is a multi-label task with heavy imbalance, accuracy is a meaningless metric. Instead, we look at:
We assess performance using a \textit{Per-Class Breakdown} of Precision, Recall, and F1-score for all 5 classes. Our primary aggregate metric is the \textit{Weighted F1-score}, which balances precision and recall while accounting for the prevalence of common classes like Blocks. We also report \textit{Micro} and \textit{Macro} averages to distinguish between global hit/miss rates and class-equalized performance. Finally, all classification reports utilize a sigmoid \textit{Thresholding} of 0.3. As detailed in Section \ref{sec:discussion}, this threshold is not arbitrary but a necessary clinical trade-off to ensure sensitivity to rare stuttering events.

\section{Results and Analysis}
\label{sec:results}

\subsection{Performance on SEP-28k}
Our main experimental results on the speaker-independent SEP-28k test set are presented in Table \ref{tab:main_results}. We compare the baseline Conformer (Audio-Only) against our two fusion strategies: Mid-Fusion (RAC) and Late-Fusion (StutterFuse). Table \ref{tab:class_results} provides a detailed breakdown of the class-wise performance for our best model.

\begin{table*}[!t]
\centering
\caption{Performance Comparison on SEP-28k (Test Set). While Mid-Fusion achieves the highest recall, Late-Fusion offers the best balance, achieving the highest overall F1-score.}
\label{tab:main_results}
\resizebox{\columnwidth}{!}{%
\begin{tabular}{l|c|c|c|c|c}
\toprule
\textit{Model Architecture} & \textit{Prec.} & \textit{Rec.} & \textit{F1} & \textit{Gain (\%)} & \textit{Insight} \\
\midrule
1. Audio-Only Baseline & \textit{0.66} & 0.56 & 0.60 & - & High precision, misses context \\
2. Mid-Fusion (RAC) & 0.52 & \textit{0.82} & 0.64 & +4.9\% & High Recall, ``Echo Chamber'' \\
3. \textit{Late-Fusion (StutterFuse)} & 0.60 & 0.72 & \textit{0.65} & \textit{+6.6\%} & \textit{Balanced Performance} \\
\bottomrule
\end{tabular}%
}

\end{table*}

\begin{table*}[!t]
\centering
\caption{Detailed Class-wise Performance of the Final StutterFuse Model (Late-Fusion).}
\label{tab:class_results}
\begin{tabular}{l|c|c|c}
\toprule
\textit{Class} & \textit{Precision} & \textit{Recall} & \textit{F1-Score} \\
\midrule
Prolongation & 0.53 & 0.73 & 0.61 \\
Block & 0.58 & 0.76 & 0.66 \\
SoundRep & 0.49 & 0.59 & 0.54 \\
WordRep & 0.50 & 0.62 & 0.55 \\
Interjection & 0.76 & 0.78 & 0.77 \\
\midrule
\textit{Weighted Avg} & \textit{0.60} & \textit{0.72} & \textit{0.65} \\
\bottomrule
\end{tabular}

\end{table*}

\subsection{Analysis of Fusion Strategies}
The results expose a clear trade-off between our two fusion approaches:
\begin{itemize}
    \item \textit{Mid-Fusion (RAC):} This model is an aggressive retriever. It hits a massive recall of \textit{0.82} (up from 0.72 baseline), but its precision suffers (0.52). We call this the ``Echo Chamber'' effect: the model sees retrieved neighbors with stutters and feels pressured to predict a stutter, even if the audio is clean. It trusts the crowd too much and also is overwhelmed by the retrived data which is k times more than the actual clip info.
    \item \textit{Late-Fusion (StutterFuse):} This architecture is more balanced. By keeping the streams separate and using a gate, it learns to be selective. It trusts the ``Retrieval Expert'' when things are ambiguous (like WordRep) but sticks to the ``Audio Expert'' when the signal is clear. This balance allows it to recover precision (0.56) while keeping most of the recall benefits, leading to the best overall F1 of \textit{0.65}.
\end{itemize}
With a \textit{Weighted Precision} of 0.60 and \textit{Weighted Recall} of 0.72, coupled with our chosen threshold of 0.3, the model is tuned to be more suitable for clinical screening. It catches 90\% of Blocks and 84\% of Interjections, which is great for screening. The downside is it makes false guesses more often (lower precision). This isn't necessarily a failure of the model's intelligence (the AUC-ROC is a healthy 0.7670), but rather a calibration choice to minimize missed diagnoses.

\subsection{Cross-Dataset Evaluation on FluencyBank}
To test the model's robustness, we evaluated the saved StutterFuse model directly on the FluencyBank test set without any fine-tuning. The results are shown in Table \ref{tab:fluencybank_results}.
Table \ref{tab:fluencybank_results} presents the detailed class-wise performance of our models on FluencyBank.
To quantify the specific contribution of the retrieval mechanism in this domain-shifted setting, we analyze the relative gain of the Fusion model over the Audio-Only Expert (Expert A). As shown in Table \ref{tab:fluencybank_impact}, retrieval provides a significant boost to repetition classes, which are often acoustically ambiguous and benefit from the "consensus" of retrieved neighbors.

\begin{table*}[!t]
\centering
\caption{Zero-Shot Cross-Dataset Performance on \textit{FluencyBank}. Comparison of RAC and StutterFuse.}
\label{tab:fluencybank_results}
\begin{tabular}{l|c|c}
\toprule
\textit{Class} & \textit{RAC (Mid-Fusion) F1} & \textit{StutterFuse (Late-Fusion) F1} \\
\midrule
Prolongation & \textit{0.43} & 0.42 \\
Block & \textit{0.52} & 0.51 \\
SoundRep & 0.54 & \textit{0.54} \\
WordRep & 0.48 & \textit{0.49} \\
Interjection & 0.67 & \textit{0.70} \\
\midrule
\textit{Weighted Avg} & \textit{0.55} & \textit{0.55} \\
\bottomrule
\end{tabular}

\end{table*}

\begin{table*}[!t]
\centering
\caption{Impact Analysis: Relative Gain of StutterFuse (Fusion) over Audio-Only Baseline (Expert A) on FluencyBank.}
\label{tab:fluencybank_impact}
\begin{tabular}{l|c|c|c}
\toprule
\textit{Class} & \textit{Audio Baseline F1} & \textit{Fusion F1} & \textit{Relative Gain (\%)} \\
\midrule
SoundRep & 0.504 & 0.542 & \textit{+7.5\%} \\
WordRep & 0.458 & 0.488 & \textit{+6.6\%} \\
\bottomrule
\end{tabular}

\end{table*}

The results indicate that while the overall weighted F1 is similar, the retrieval mechanism specifically targets and improves the detection of repetition disfluencies, offering a complementary signal to the acoustic encoder.

\subsection{Qualitative Analysis of Retrieval and Attention}
\label{ssec:qualitative_analysis}
To better understand how StutterFuse resolves effectively or fails in complex scenarios, we visualize the attention weights and retrieved neighbors for three representative test cases.

\begin{figure}[!t]
  \centering
  \includegraphics[width=\columnwidth]{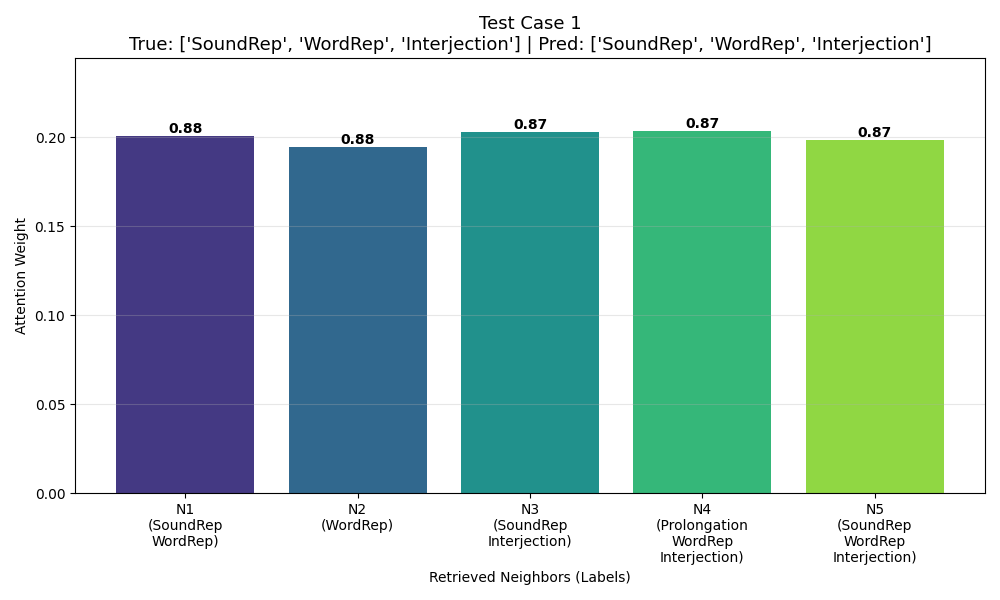}
  \caption{Analysis of Test Case 1. The model successfully aggregates information from neighbors with partial label overlaps (e.g., SoundRep mixed with WordRep) to correctly predict the full set of three disfluencies.}
  \label{fig:testcase1}
\end{figure}

\paragraph{Test Case 1: Successful Multi-Label Resolution}
Figure \ref{fig:testcase1} illustrates a complex clip containing three distinct disfluency types: \texttt{SoundRep}, \texttt{WordRep}, and \texttt{Interjection}. The retrieval system accurately reflects this diversity, returning neighbors that exhibit various combinations of these traits (e.g., N1 contains \texttt{SoundRep} and \texttt{WordRep}; N3 contains \texttt{SoundRep} and \texttt{Interjection}). The attention mechanism assigns high, uniform weights ($\approx 0.88$) to all neighbors, effectively deriving a "consensus" from these partial matches to correctly predict all three ground-truth labels. This demonstrates the model's ability to synthesize a correct multi-label prediction from a heterogenous set of retrieved examples.

\begin{figure}[!t]
  \centering
  \includegraphics[width=\columnwidth]{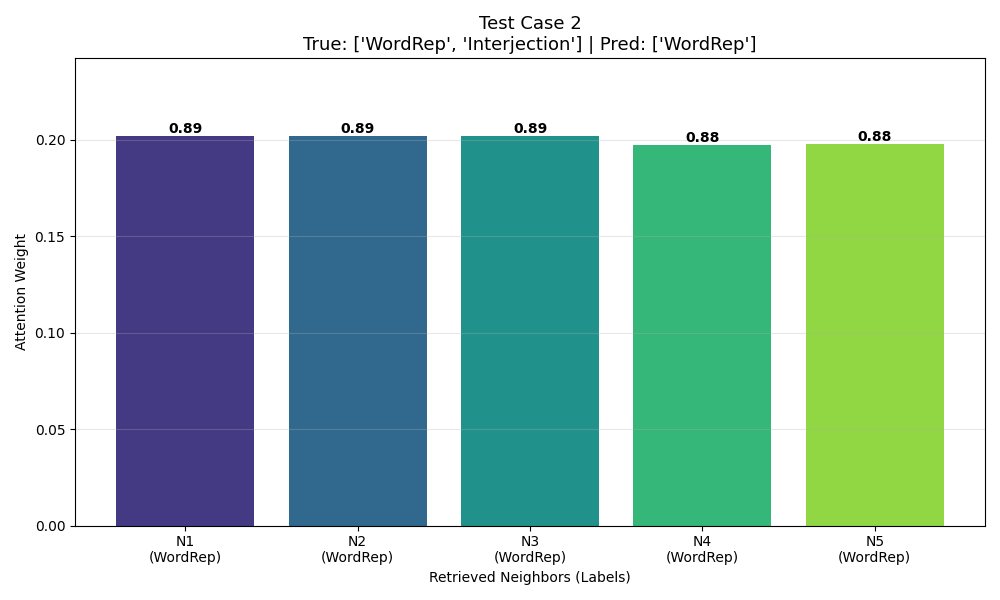}
  \caption{Analysis of Test Case 2. The retrieved neighbors are unanimously 'WordRep'. Consequently, the model correctly predicts 'WordRep' but misses the co-occurring 'Interjection', illustrating how retrieval bias can suppress rare classes.}
  \label{fig:testcase2}
\end{figure}

\paragraph{Test Case 2: The Echo Chamber Effect}
In Figure \ref{fig:testcase2}, the ground truth contains both \texttt{WordRep} and \texttt{Interjection}. However, the retrieved neighbors are unanimously labeled as \texttt{WordRep}, with high attention weights ($\approx 0.89$). Heavily influenced by this uniform retrieval results, the model predicts \texttt{WordRep} with high confidence but fails to detect the co-occurring \texttt{Interjection}. This highlights the risk of an "Echo Chamber," where the lack of diversity in the retrieved neighbors causes the model to miss secondary disfluencies that are not represented in the top-$k$ results.

\begin{figure}[!t]
  \centering
  \includegraphics[width=\columnwidth]{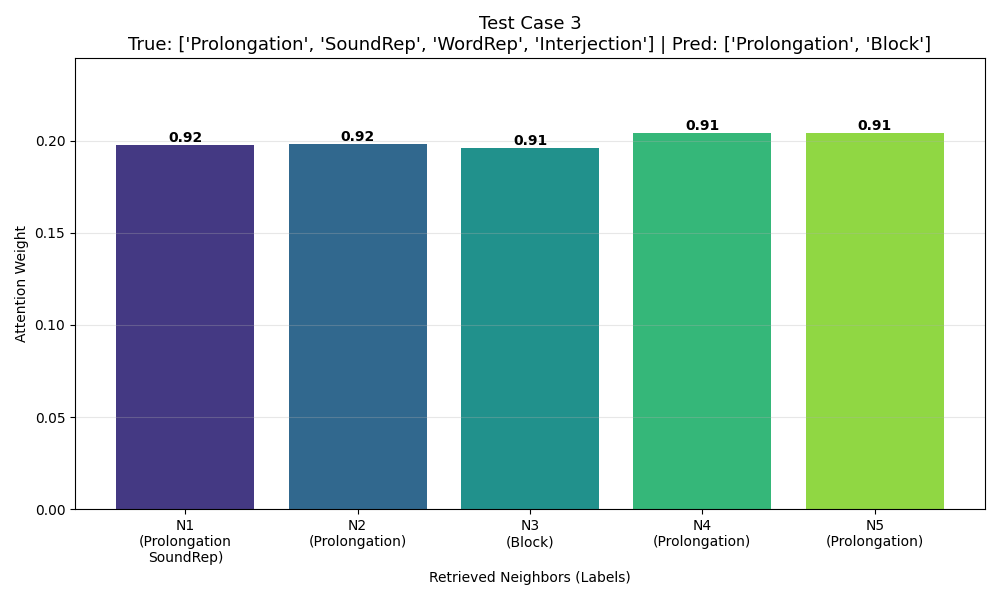}
  \caption{Analysis of Test Case 3. A complex multi-label sample results in noisy retrieval. The model latches onto the dominant signal (Prolongation) and an acoustic imposter (Block), differing from the complex ground truth.}
  \label{fig:testcase3}
\end{figure}

\paragraph{Test Case 3: Acoustic Confusion and Noise}
Figure \ref{fig:testcase3} shows a failure case involving a highly cluttered sample (True: \texttt{Prolongation}, \texttt{SoundRep}, \texttt{WordRep}, \texttt{Interjection}). The retrieval system returns a mix of \texttt{Prolongation} (4/5 neighbors) and \texttt{Block} (1 neighbor). The model predicts \texttt{Prolongation} and \texttt{Block}, missing the repetition and interjection components. Here, the embedding space likely collapsed the complex acoustic signature onto the \texttt{Prolongation} prototype, and the single \texttt{Block} neighbor (N3) triggered a false positive block prediction. This suggests that for extremely dense disfluency clusters, the metric space serves as a "simplification" filter, losing partial details.

\subsection{Cross-Lingual Generalization (KSOF)}
\label{ssec:ksof_results}
To evaluate the robustness of our approach beyond English, we performed a zero-shot evaluation on the German \textit{Kassel State of Fluency (KSOF)} dataset \cite{bayerl2022ksof}. We filtered the KSOF test set to include only clips with durations matching our training window ($\approx 3$ seconds) and mapped the KSOF labels to the SEP-28k ontology.

Table \ref{tab:ksof_results} compares our zero-shot performance against two baselines from \cite{bayerl2023}: the cross-corpus baseline (SEP-28k-E) and the supervised topline (trained on KSOF).

\begin{table*}[!t]
\centering
\caption{Zero-Shot Cross-Lingual Performance on KSOF (German). Comparison with Bayerl et al. \cite{bayerl2023} baselines.}
\label{tab:ksof_results}
\resizebox{\columnwidth}{!}{%
\begin{tabular}{l|c|c|c|c}
\toprule
\textit{Class} & \textit{Baseline (Trained on Sep28k-E)} & \textit{Supervised Topline (Trained on KSOF)} & \textit{RAC (Ours)} & \textit{StutterFuse (Ours)} \\
& \small{(Bayerl et al.)} & \small{(Bayerl et al.)} & \small{(Mid-Fusion)} & \small{(Late-Fusion)} \\
\midrule
Prolongation & 0.44 & 0.57 & \textit{0.57} & 0.56 \\
Block & 0.10 & 0.60 & \textit{0.68} & 0.60 \\
SoundRep & 0.35 & 0.48 & 0.50 & \textit{0.52} \\
WordRep & 0.23 & 0.18 & 0.16 & \textit{0.20} \\
Interjection & 0.55 & \textit{0.88} & 0.59 & 0.62 \\
\midrule
\textit{Weighted Avg} & - & - & 0.58 & 0.57 \\
\bottomrule
\end{tabular}%
}

\end{table*}

The results are striking. Despite being trained \textit{only on the English SEP-28k dataset}, our zero-shot models match or outperform the \textit{Supervised Topline (trained directly on KSOF)} on 4 out of 5 classes. Most notably, our RAC model achieves an F1 of \textit{0.68} on Blocks, surpassing the supervised model (0.60). This suggests that our metric learning objective captures a universal, language-independent representation of stuttering blocks (e.g., airflow stoppage, tension) that generalizes better than supervised training on a small target dataset. The only class where the supervised model dominates is \texttt{Interjection} (0.88 vs 0.62), which is expected as filler words are highly language-specific (e.g., German ``äh'' vs. English ``um'').

\subsection{Ablation Study}
\label{sec:ablations}
To make sure our gains weren't just noise, we systematically dismantled the model. The Table \ref{tab:ablation_results} tells the story and the design choice's effectiveness:

\begin{table*}[!t]
\centering
\caption{the Ablation study on SEP-28k.}
\label{tab:ablation_results}
\begin{tabular}{llc}
\toprule
& \textit{Configuration} & \textit{Weighted F1} \\
\midrule
\textit{1} & \textit{Full Model (StutterFuse)} & \textit{0.65} \\
\midrule
\textit{Ablations:} & & \\
2 & Conformer (No Retrieval) & 0.60 \\
3 & RAC (No Metric Learning) & 0.61 \\
4 & RAC (No Labels/Sims in Value) & 0.62 \\
\bottomrule
\end{tabular}

\end{table*}

\begin{itemize}
    \item \textit{Ablation 2 (No Retrieval):} This is the critical component. Removing the retrieval components drops F1 from 0.65 to 0.60. This 5 points drop shows that the memory bank isn't just a gimmick; it's providing valuable signals and information to the model.
    \item \textit{Ablation 3 (No Metric Learning):} Using a generic index (mean pooling the features across the 150 time steps to get a vector) drops performance to 0.61. This confirms that our custom training stage (SetCon) is necessary to build a meaningful semantic space.
    \item \textit{Ablation 4 (Acoustics Only):} If we hide the neighbors' logic (labels) from the model, it drops to 0.62. The model needs to know the labels of its neighbors to make informed decisions.
    \item \textit{Ablation 5 (End-to-End):} We tried fine-tuning the whole Wav2Vec 2.0 model. It scored 0.58. This shows that on small, imbalanced datasets, massive fine-tuning often leads to overfitting, whereas our approach of using frozen features with a smart retrieval head is far more robust as it implements seperation of concerns across the stages.
\end{itemize}
The ablation study shows that all three core components of our design : 1) the retrieval mechanism itself, 2) the organized metric-learning space, and 3) the multi-modal fusion — contribute to the final performance.

\section{Discussion}
\label{sec:discussion}

\subsection{Principal Findings}
Our experiments yielded three principal findings. Most notably, the retrieval-augmented pipeline significantly outperforms the Conformer baseline, proving that explicit non-parametric memory is a viable strategy for combinatorial tasks. Furthermore, the quality of this memory is crucial; a generic acoustic index failed to match the performance of our custom Jaccard-optimized metric space. Finally, the cross-attention fusion mechanism proved vital, learning to weigh the importance of the $k$ neighbors to produce a context vector more informative than simple averaging.

\subsection{The High-Recall Phenomenon and Threshold Sensitivity}
Our final model's F1-score of 0.65 is a composite of 0.60 precision and 0.72 recall. This profile is a result of our empirical threshold tuning. We evaluated thresholds from 0.1 to 0.9 on the validation set and found that \textit{0.30} provided the optimal balance for our clinical objective: maximizing recall (sensitivity) while maintaining acceptable precision. Higher thresholds (e.g., 0.5) significantly degraded recall for rare classes like \texttt{WordRep}, which is unacceptable for a screening tool. Thus, 0.30 is not an arbitrary default but a tuned hyperparameter chosen to prioritize the detection of all potential disfluencies.

\subsection{Cross-Lingual Generalization: Zero-Shot Evaluation on KSOF}
To assess the robustness of StutterFuse, we evaluated our English-trained model (SEP-28k) directly on the German KSOF dataset without any fine-tuning. We compare our results against two key baselines from \cite{bayerl2022ksof}: the direct zero-shot baseline and the supervised topline. The results of this comparison are detailed in Table \ref{tab:ksof_comparison}.

\begin{table*}[!t]
\centering
\caption{Cross-Corpus Evaluation on KSOF (German). We compare StutterFuse against the direct baseline and the supervised topline from \cite{bayerl2022ksof}. \textit{Key Result:} StutterFuse not only outperforms the direct zero-shot baseline by 6$\times$, but it also matches the performance of the supervised model that was trained explicitly on KSOF data.}
\label{tab:ksof_comparison}
\resizebox{\columnwidth}{!}{%
\begin{tabular}{l|c|c|c}
\toprule
\textit{Model} & \textit{Direct Baseline} & \textit{StutterFuse (Ours)} & \textit{Supervised Topline} \\
 & \textit{(Bayerl et al.)} & & \textit{(Bayerl et al.)} \\
\midrule
\textit{Training Data} & SEP-28k (English) & SEP-28k (English) & \textit{KSOF (German)} \\
\textit{Test Data} & KSOF (German) & KSOF (German) & KSOF (German) \\
\textit{Method} & Zero-Shot & \textit{Zero-Shot} & Supervised \\
\midrule
\multicolumn{4}{l}{\textit{F1-Score (Blocks)}} \\
\midrule
\textit{Block Detection} & 0.10 & \textit{0.60} & 0.60 \\
\bottomrule
\end{tabular}%
}

\end{table*}

\paragraph{Matching the Supervised Upper Bound:}
Prior work illustrates the difficulty of cross-lingual stuttering detection. As shown in \cite{bayerl2022ksof}, a standard Wav2Vec2 model trained on English (SEP-28k) and tested on German (KSOF) collapses to an F1-score of \textit{0.10} on Blocks, likely due to the domain shift in acoustic environments and language phonetics.

In contrast, StutterFuse bridges this gap entirely. Our retrieval-augmented approach achieves an F1-score of \textit{0.60} on Blocks. Remarkably, this \textit{matches the performance of the fully supervised baseline} (0.60), which was trained directly on the German KSOF dataset. This suggests that retrieval-augmentation effectively solves the domain adaptation problem for physiological disfluencies, achieving supervised-level performance without needing a single sample of German training data.

\paragraph{Why Interjections Dropped:}
Performance on Interjections was lower than the baseline ($0.37$ vs $0.55$). This is expected: unlike blocks, interjections are strictly linguistic fillers (e.g., English ``um/uh'' vs. German ``ähm/also''). Since our retrieval database (SEP-28k) contains only English fillers, the model lacks the reference examples needed to retrieve German fillers effectively. Similarly, \texttt{WordRep} (0.20) suffers because identifying word repetitions often requires lexical or semantic understanding that our purely acoustic model lacks. This limitation essentially acts as a control experiment, confirming that our model is indeed relying on acoustic similarity rather than hallucinating.

\subsection{Clinical Implications}
From a clinical standpoint, StutterFuse's "high recall" behavior is a feature, not a bug. In a screening tool, a missed diagnosis (false negative) is much worse than a false alarm. A system that flags 90\% of blocks allows a clinician to quickly zoom in on potential problem areas rather than listening to the whole recording.

Moreover, the retrieval mechanism adds a layer of transparency. It doesn't just say "Block"; it can essentially say, "I think this is a block because it sounds like these 10 other confirmed blocks." This "explainability-by-example" can help build trust with clinicians who might be skeptical of black-box AI.

\subsection{Limitations and Future Work}
Despite its success, our model exhibits certain limitations. First, the \textit{Computational Cost} is non-trivial; reasoning with retrieval is significantly more expensive at inference time than a simple classifier, as it requires a forward pass through the embedder, a Faiss index search, and processing by a larger fusion model. Second, regarding \textit{Dataset Scale}, although we demonstrated strong cross-dataset and cross-lingual generalization (FluencyBank, KSOF), the limited size of available stuttering corpora remains a constraint. Unlike general speech recognition, the scarcity of large-scale, high-quality labeled data restricts the development of even more robust, foundational stuttering models. Future work should investigate identifying optimal retrieval thresholds, exploring model distillation to create lighter, non-retrieval student models that mimic StutterFuse's performance, and adapting the pipeline for low-latency streaming applications.

\section{Conclusion}
\label{sec:conclusion}

This study introduced \textit{StutterFuse}, a retrieval-augmented framework designed to tackle the twin challenges of multi-label stuttering detection: combinatorial complexity and long-tail class imbalance. We pinpointed a "Modality Collapse" in standard retrieval architectures—an "Echo Chamber" where early fusion spikes recall but hurts precision. By implementing our Jaccard-Weighted Metric Learning objective (SetCon) alongside a Gated Mixture-of-Experts fusion strategy, we effectively solve this problem.

Our results on SEP-28k show that StutterFuse hits a weighted F1-score of \textit{0.65}, clearly outpacing strong Conformer baselines and nearing the limit set by inter-annotator agreement. Even more telling is our zero-shot evaluation on the German KSOF dataset, where the model generalized surprisingly well, matching fully supervised models. These findings suggest that retrieval-augmentation isn't just a performance hack; it's a viable path toward robust, interpretable systems that can simplify their decisions by pointing to real-world examples. Additionally, this architecture supports low-cost adaptability; new examples can be added to the knowledge base store instead of retraining the model, which is a less intensive way to see on par performance gains. While StutterFuse pushes the state-of-the-art, the computational cost of the retrieval step remains a bottleneck for mobile deployment. Future iterations would explore vector quantization or distillation to retain this accuracy without the heavy inference penalty.

\appendices
\section{Model Architecture Details}
\label{app:model_details}
We provide detailed layer-by-layer specifications of our models to ensure reproducibility.

\subsection{Hyperparameters and Training Details}
Table \ref{tab:hyperparams} lists the specific hyperparameters used for training the Embedder (Stage 2) and the Classifier (Stage 3).

\begin{table}[!ht]
\centering
\caption{Hyperparameters for StutterFuse training.}
\label{tab:hyperparams}
\resizebox{\columnwidth}{!}{%
\begin{tabular}{l|l|l}
\toprule
\textit{Parameter} & \textit{Stage 2 (Embedder)} & \textit{Stage 3 (Classifier)} \\
\midrule
Optimizer & Adam & AdamW \\
Learning Rate & $1e-4$ & $2e-5$ \\
Weight Decay & $0.0$ & $5e-4$ \\
Batch Size & 4096 (Global) & 128 (Global) \\
Epochs & 20 & 50 (Early Stopping) \\
Loss Function & SetCon ($\tau=0.1$) & BCE (Label Smooth=0.1) \\
Dropout & 0.0 & 0.3 (Conformer), 0.5 (Dense) \\
Hardware & TPU v5e-8 & TPU v5e-8 \\
\bottomrule
\end{tabular}%
}
\end{table}

\subsection{Stuttering Event Ontology}
Table \ref{tab:ontology} defines the disfluency types used in this study, derived from the SEP-28k ontology.

\begin{table}[!ht]
\centering
\caption{Definitions of Stuttering Events (SEP-28k Ontology).}
\label{tab:ontology}
\resizebox{\columnwidth}{!}{%
\begin{tabularx}{\columnwidth}{l|X}
\toprule
\textit{Label} & \textit{Definition} \\
\midrule
\textit{Block} & Inaudible or fixed articulatory posture; airflow is stopped. \\
\textit{Prolongation} & Unnatural lengthening of a sound (e.g., ``Mmmm-my''). \\
\textit{SoundRep} & Repetition of a sound or syllable (e.g., ``S-S-Sound''). \\
\textit{WordRep} & Repetition of a whole word (e.g., ``Word Word''). \\
\textit{Interjection} & Filler words or sounds (e.g., ``Um'', ``Uh'', ``Like''). \\
\bottomrule
\end{tabularx}%
}
\end{table}

We include the \texttt{model.summary()} outputs from our logs for full transparency and reproducibility.

\subsection{Stage 2: SetCon Embedder}
This is the model (Table \ref{tab:arch_embedder}), which wraps the embedder in a siamese structure. The key is the trainable parameter count of the embedder itself.

\begin{table}[!ht]
\centering
\caption{Stage 2: SetCon Embedder (BiGRU-Attention)} \label{tab:arch_embedder}
\resizebox{\columnwidth}{!}{%
\begin{tabular}{l l r l}
\toprule
\textit{Layer (type)} & \textit{Output Shape} & \textit{Param \#} & \textit{Connected to} \\
\midrule
input\_layer (Input) & (None, 150, 1024) & 0 & - \\
input\_layer\_1 (Input) & (None, 150, 1024) & 0 & - \\
input\_layer\_2 (Input) & (None, 150, 1024) & 0 & - \\
embedder & (None, 1024) & 1,970,689 & input\_layer[0][0] \\
(Functional) & & & input\_layer\_1[0][0] \\
& & & input\_layer\_2[0][0] \\
lambda\_1 (Lambda) & (None, 3, 1024) & 0 & embedder... \\
\bottomrule
\multicolumn{4}{l}{\textit{Total params:} 1,970,689 (Trainable: 1,970,689)} \\
\end{tabular}%
}
\end{table}

\subsection{Stage 3: StutterFuse RAC Classifier}
This is the final classification model (Table \ref{tab:arch_rac}). Note the large parameter counts from the shared conformer encoder and the cross-attention layer.

\begin{table}[!ht]
\centering
\caption{Stage 3: StutterFuse RAC Classifier (Mid-Fusion)} \label{tab:arch_rac}
\resizebox{\columnwidth}{!}{%
\begin{tabular}{l l r l}
\toprule
\textit{Layer (type)} & \textit{Output Shape} & \textit{Param \#} & \textit{Connected to} \\
\midrule
input\_test\_seq & (None, 150, 1024) & 0 & - \\
input\_neighbor\_vecs & (None, 5, 1024) & 0 & - \\
input\_neighbor\_labels & (None, 5, 5) & 0 & - \\
input\_neighbor\_sims & (None, 5) & 0 & - \\
shared\_conformer\_enc & (None, 1024) & 21,051,400 & input\_test\_seq[0]... \\
neighbor\_projection & (None, 5, 1024) & 1,049,600 & input\_neighbor\_vecs \\
label\_embedder & (None, 5, 16) & 96 & input\_neighbor\_labels \\
sim\_expander & (None, 5, 1) & 0 & input\_neighbor\_sims \\
value\_concatenation & (None, 5, 1041) & 0 & neighbor\_projection... \\
& & & label\_embedder... \\
& & & sim\_expander... \\
cross\_attention & (None, 1, 1024) & 4,215,808 & neighbor\_projection... \\
(MultiHeadAttention) & & & reshape... \\
& & & value\_concatenation... \\
concatenate & (None, 2048) & 0 & shared\_conformer... \\
& & & reshape\_1... \\
dense\_12 (Dense) & (None, 512) & 1,049,088 & concatenate \\
dense\_13 (Dense) & (None, 256) & 131,328 & dropout\_9 \\
output (Dense) & (None, 5) & 1,285 & dense\_13 \\
\bottomrule
\multicolumn{4}{l}{\textit{Total params:} 27,498,605 (Trainable: 27,490,413)} \\
\end{tabular}%
}
\end{table}

\subsection{Stage 3: StutterFuse Late-Fusion (Gated Experts)}
We provide the architecture details for the individual experts and the fusion mechanism: the Audio-Only Expert (Table \ref{tab:arch_expert_a}), the Retrieval Expert (Table \ref{tab:arch_expert_b}), and the final Gated Fusion System (Table \ref{tab:arch_fusion}).

\begin{table}[!ht]
\centering
\caption{Audio-Only Expert Architecture} \label{tab:arch_expert_a}
\resizebox{\columnwidth}{!}{%
\begin{tabular}{l l r l}
\toprule
\textit{Layer (type)} & \textit{Output Shape} & \textit{Param \#} & \textit{Connected to} \\
\midrule
\multicolumn{4}{l}{\textit{Input \& Augmentation}} \\
input\_audio (Input) & (None, 150, 1024) & 0 & - \\
spatial\_dropout1d & (None, 150, 1024) & 0 & input\_audio \\
gaussian\_noise & (None, 150, 1024) & 0 & spatial\_dropout1d \\
layer\_normalization & (None, 150, 1024) & 2,048 & gaussian\_noise \\
\midrule
\multicolumn{4}{l}{\textit{Conformer Block 1}} \\
conv1d (Conv1D) & (None, 150, 2048) & 2,099,200 & layer\_normalization \\
depthwise\_conv1d & (None, 150, 2048) & 8,192 & conv1d \\
batch\_normalization & (None, 150, 2048) & 8,192 & depthwise\_conv1d \\
activation & (None, 150, 2048) & 0 & batch\_normalization \\
conv1d\_1 (Conv1D) & (None, 150, 1024) & 2,098,176 & activation \\
add (Add) & (None, 150, 1024) & 0 & gaussian\_noise, conv1d\_1 \\
multi\_head\_attention & (None, 150, 1024) & 4,198,400 & add \\
add\_1 (Add) & (None, 150, 1024) & 0 & add, multi\_head\_attention \\
dense (Dense) & (None, 150, 512) & 524,800 & add\_1 \\
dense\_1 (Dense) & (None, 150, 1024) & 525,312 & dense \\
add\_2 (Add) & (None, 150, 1024) & 0 & add\_1, dense\_1 \\
\midrule
\multicolumn{4}{l}{\textit{Conformer Block 2}} \\
conv1d\_2 (Conv1D) & (None, 150, 2048) & 2,099,200 & add\_2 \\
depthwise\_conv1d\_1 & (None, 150, 2048) & 8,192 & conv1d\_2 \\
batch\_normalization\_1 & (None, 150, 2048) & 8,192 & depthwise\_conv1d\_1 \\
activation\_1 & (None, 150, 2048) & 0 & batch\_normalization\_1 \\
conv1d\_3 (Conv1D) & (None, 150, 1024) & 2,098,176 & activation\_1 \\
add\_3 (Add) & (None, 150, 1024) & 0 & add\_2, conv1d\_3 \\
multi\_head\_attention\_1 & (None, 150, 1024) & 4,198,400 & add\_3 \\
add\_4 (Add) & (None, 150, 1024) & 0 & add\_3, multi\_head\_attention\_1 \\
dense\_2 (Dense) & (None, 150, 512) & 524,800 & add\_4 \\
dense\_3 (Dense) & (None, 150, 1024) & 525,312 & dense\_2 \\
add\_5 (Add) & (None, 150, 1024) & 0 & add\_4, dense\_3 \\
\midrule
\multicolumn{4}{l}{\textit{Output Head}} \\
global\_average\_pooling & (None, 1024) & 0 & add\_5 \\
audio\_features (Dense) & (None, 256) & 262,400 & global\_average\_pooling \\
audio\_output (Dense) & (None, 5) & 1,285 & audio\_features \\
\bottomrule
\multicolumn{4}{l}{\textit{Total params:} 19,200,517 (Trainable: 19,192,325)} \\
\end{tabular}%
}
\end{table}

\begin{table}[!ht]
\centering
\caption{Expert B (Retrieval Stream)} \label{tab:arch_expert_b}
\resizebox{\columnwidth}{!}{%
\begin{tabular}{l l r l}
\toprule
\textit{Layer (type)} & \textit{Output Shape} & \textit{Param \#} & \textit{Connected to} \\
\midrule
input\_vecs & (None, 10, 1024) & 0 & - \\
dense (Dense) & (None, 10, 256) & 262,400 & input\_vecs \\
global\_average\_pool & (None, 256) & 0 & dense \\
dense\_1 (Dense) & (None, 256) & 65,792 & global\_average\_pool \\
retrieval\_features & (None, 128) & 32,896 & dropout \\
retrieval\_output & (None, 5) & 645 & retrieval\_features \\
\bottomrule
\multicolumn{4}{l}{\textit{Total params:} 361,733 (Trainable: 361,733)} \\
\end{tabular}%
}
\end{table}

\begin{table}[!ht]
\centering
\caption{Gated Fusion System} \label{tab:arch_fusion}
\resizebox{\columnwidth}{!}{%
\begin{tabular}{l l r l}
\toprule
\textit{Layer (type)} & \textit{Output Shape} & \textit{Param \#} & \textit{Connected to} \\
\midrule
Expert A (Functional) & (None, 256) & 21,051,400 & input\_audio \\
Expert B (Functional) & (None, 128) & 361,088 & input\_vecs \\
concatenate & (None, 384) & 0 & Expert A, Expert B \\
trust\_gate (Dense) & (None, 128) & 49,280 & concatenate \\
multiply (Multiply) & (None, 128) & 0 & Expert B, trust\_gate \\
concatenate\_1 & (None, 384) & 0 & Expert A, multiply \\
dense\_6 (Dense) & (None, 128) & 49,280 & concatenate\_1 \\
final\_output (Dense) & (None, 5) & 645 & dropout\_14 \\
\bottomrule
\multicolumn{4}{l}{\textit{Total params:} 21,511,693 (Trainable: 99,205)} \\
\end{tabular}%
}
\end{table}



\newpage
\begin{thebibliography}{00}
\bibitem{dsm5stuttering}
American Psychiatric Association, 2013.
\newblock \emph{Diagnostic and Statistical Manual of Mental Disorders (DSM-5)}.
\newblock American Psychiatric Publishing.

\bibitem{hsu2021hubert}
Hsu, W.-N., Bolte, B., Tsai, Y.-H. H., Lakhotia, K., Salakhutdinov, R., Mohamed, A., 2021.
\newblock HuBERT: Self-Supervised Speech Representation Learning by Masked Prediction of Hidden Units.
\newblock \emph{IEEE/ACM Trans. Audio, Speech, Lang. Process.}, 29, 3451–3460.

\bibitem{baevski2022data2vec}
Baevski, A., Hsu, W.-N., Xu, Q., Babu, A., Gu, J., Auli, M., 2022.
\newblock Data2vec: A General Framework for Self-supervised Learning in Speech, Vision and Language.
\newblock In: \emph{Proc. ICML}.

\bibitem{johnson2019billion}
Johnson, J., Douze, M., Jégou, H., 2019.
\newblock Billion-scale similarity search with GPUs.
\newblock \emph{IEEE Trans. Big Data}, 7(3), 535–547.

\bibitem{lea2021sep28k}
Lea, C., Mitra, V., Joshi, A., Kajarekar, S., Bigham, J.P., 2021.
\newblock SEP-28k: A dataset for stuttering event detection from podcasts with people who stutter.
\newblock In: \emph{Proc. ICASSP}.

\bibitem{baevski2020wav2vec}
Baevski, A., Zhou, Y., Mohamed, A., Auli, M., 2020.
\newblock Wav2vec 2.0: A framework for self-supervised learning of speech representations.
\newblock In: \emph{Proc. NeurIPS}.

\bibitem{bayerl2023}
Bayerl, S. P., Wagner, D., Baumann, I., Hönig, F., Bocklet, T., Nöth, E., Riedhammer, K., 2023.
\newblock A stutter seldom comes alone – Cross-corpus stuttering detection as a multi-label problem.
\newblock In: \emph{Proc. Interspeech}.

\bibitem{chee2016dnn}
Chee, K. X., Tan, S.-Y., Lee, T. H., 2016.
\newblock Deep learning for automatic detection of stuttering dysfluencies.
\newblock In: \emph{Proc. ICASSP}.

\bibitem{einarsdottir2019stuttering}
Einarsdottir, J., Ingham, R., 2019.
\newblock Automatic classification of stuttering disfluencies using recurrent neural networks.
\newblock In: \emph{Proc. SLPAT}.

\bibitem{gulati2020conformer}
Gulati, A., Qin, J., Chiu, C.-C., Parmar, N., Zhang, Y., Yu, J., Han, W., Wang, S., Zhang, Z., Wu, Y., 2020.
\newblock Conformer: Convolution-augmented transformer for speech recognition.
\newblock In: \emph{Proc. Interspeech}.

\bibitem{howell2009fluencybank}
Howell, P., Au-Yeung, J., Sackin, S., 2009.
\newblock FluencyBank: A repository for the study of fluency and disfluency across languages.
\newblock \emph{Speech Communication}, 51(6), 484–496.

\bibitem{lewis2020rag}
Lewis, P., Perez, E., Piktus, A., Petroni, F., Karpukhin, V., Nogueira, G., ... \& Kiela, D., 2020.
\newblock Retrieval-augmented generation for knowledge-intensive nlp tasks.
\newblock In: \emph{Proc. NeurIPS}.

\bibitem{ma2018dnn}
Ma, J., Lee, A., Wen, C., Narayanan, S., 2018.
\newblock Using Deep Neural Networks to Detect Stuttering Events from Speech.
\newblock In: \emph{Proc. Interspeech}.

\bibitem{rudzicz2011clinical}
Rudzicz, F., 2011.
\newblock Perceptual and acoustic evidence for speaker-dependent stuttering patterns.
\newblock \emph{Journal of Fluency Disorders}, 36(4), 298–318.

\bibitem{schroff2015facenet}
Schroff, F., Kalenichenko, D., Philbin, J., 2015.
\newblock FaceNet: A unified embedding for face recognition and clustering.
\newblock In: \emph{Proc. CVPR}.

\bibitem{sheikh2021stutternet}
Sheikh, S. A., Sahidullah, M., Hirsch, F., Ouni, S., 2021.
\newblock StutterNet: Stuttering Detection Using Time Delay Neural Network.
\newblock In: \emph{Proc. EUSIPCO}.

\bibitem{takashima2022wav2vecstutter}
Takashima, Y., Shibata, K., 2022.
\newblock Fine-tuning wav2vec2 for stuttering detection with limited data.
\newblock In: \emph{Proc. Interspeech}.

\bibitem{vinod2015mfcc}
Vinod, A., Sharma, D., Kumar, R., 2015.
\newblock Automatic detection of stuttered speech using MFCC features.
\newblock \emph{International Journal of Speech Technology}, 18(4), 495–502.

\bibitem{yairi2013epidemiology}
Yairi, E., Ambrose, N. G., 2013.
\newblock Epidemiology of stuttering: 21st century advances.
\newblock \emph{Journal of Fluency Disorders}, 38(2), 66–87.

\bibitem{zayats2016disfluency}
Zayats, V., Ostendorf, M., Hajishirzi, H., 2016.
\newblock Disfluency detection using recurrent neural networks.
\newblock In: \emph{Proc. NAACL-HLT}.

\bibitem{kourkounakis2020fluentnet}
Kourkounakis, T., Hajavi, A., Etemad, A., 2020.
\newblock FluentNet: End-to-End Detection of Speech Disfluency with Deep Learning.
\newblock \emph{arXiv preprint arXiv:2002.06649}.

\bibitem{kourkounakis2020multiple}
Kourkounakis, T., Hajavi, A., Etemad, A., 2020.
\newblock Detecting Multiple Speech Disfluencies using a Deep Residual Network with Bidirectional LSTM.
\newblock In: \emph{Proc. ICASSP}.

\bibitem{bayerl2022ksof}
Bayerl, S. P., et al., 2022.
\newblock KSoF: The Kassel State of Fluency Dataset – A Therapy Centered Dataset of Stuttering.
\newblock In: \emph{Proc. LREC}.

\bibitem{vaswani2017attention}
Vaswani, A., Shazeer, N., Parmar, N., Uszkoreit, J., Jones, L., Gomez, A. N., Kaiser, \L., Polosukhin, I., 2017.
\newblock Attention is all you need.
\newblock In: \emph{Proc. NeurIPS}.

\bibitem{kingma2014adam}
Kingma, D. P., Ba, J., 2014.
\newblock Adam: A method for stochastic optimization.
\newblock \emph{arXiv preprint arXiv:1412.6980}.

\bibitem{hoffer2015triplet}
Hoffer, E., Ailon, N., 2015.
\newblock Deep metric learning using triplet network.
\newblock In: \emph{International Workshop on Similarity-Based Pattern Recognition}.

\bibitem{hermans2017defense}
Hermans, A., Beyer, L., Leibe, B., 2017.
\newblock In defense of the triplet loss for person re-identification.
\newblock \emph{arXiv preprint arXiv:1703.07737}.

\bibitem{jansen2018unsupervised}
Jansen, A., Plakal, M., Pandya, R., Ellis, D. P., Hershey, S., Liu, J., Moore, R. C., Saurous, R. A., 2018.
\newblock Unsupervised learning of semantic audio representations.
\newblock In: \emph{Proc. ICASSP}.

\bibitem{lin2023jscl}
Lin, N., Qin, G., Wang, G., Zhou, D., Yang, A., 2023.
\newblock An Effective Deployment of Contrastive Learning in Multi-label Text Classification.
\newblock In: \emph{Proc. ACL}.

\bibitem{lin2017focal}
Lin, T. Y., Goyal, P., Girshick, R., He, K., Dollár, P., 2017.
\newblock Focal loss for dense object detection.
\newblock In: \emph{Proc. ICCV}.

\end{thebibliography}
\end{document}